\documentclass[twocolumn]{article}

\usepackage[utf8]{inputenc}
\usepackage[english]{babel}
\usepackage{fontawesome}
\usepackage{cleveref}
\usepackage{amsmath}
\usepackage{amsthm}
\usepackage{subfig}
\usepackage{amssymb}
\usepackage{stmaryrd}
\usepackage{dsfont}
\usepackage{tikz}
\usepackage{xcolor}
\usepackage{svg}
\usepackage{soulutf8}
\usepackage{booktabs}
\usepackage{marvosym}
\usepackage{xspace}
\usepackage{pgfplots}
\usepackage{balance}
\usepackage{hyperref}
\usetikzlibrary{shapes,positioning,decorations,decorations.text,arrows,decorations.pathmorphing,patterns}

\makeatletter
\def\blfootnote{\xdef\@thefnmark{}\@footnotetext}
\makeatother

\definecolor{lviolet}{rgb}{.95,.85,1}
\definecolor{lblue}{rgb}{.85,.85,1}
\definecolor{lred}{rgb}{1,.85,.85}
\definecolor{lgreen}{rgb}{.85, 1,.85}
\definecolor{lgray}{rgb}{.92,.92,.92}
\definecolor{lorange}{rgb}{1,.92,.8}

    \title{MCTS-based Automated Negotiation Agent (Extended Abstract)}
    
    \author{Cédric L R Buron\thanks{Thales Research \& Technology, Palaiseau, France, \texttt{cedric.buron@thalesgroup.com}}, Zahia Guessoum\thanks{LIP6, Sorbonne Université, Paris, France, \texttt{zahia.guessoum@lip6.fr}}, Sylvain Ductor\thanks{Universidade Estadual do Ceará, Fortaleza, Brazil, \texttt{sylvain.ductor@uece.com.br}}}
    
\begin{document}
	\maketitle
        \blfootnote{cite as: Cédric L R Buron, Zahia Guessoum, and Sylvain Ductor. 2019. ``MCTS-based Automated Negotiation Agent (Extended Abstract)''. \textsl{In Proc. of the 18th International Conference on Autonomous Agents and Multiagent Systems (AAMAS 2019)}, Montreal, Canada, May 13–17, 2019, IFAAMAS}
        
	\begin{abstract}
	
		This paper introduces a new Negotiating Agent for automated negotiation  on continuous domains and without considering a specified deadline. The agent bidding strategy relies on Monte Carlo Tree Search, which is a trendy method since it has been used with success on games with high branching factor such as Go. It uses two opponent modeling techniques for its bidding strategy and its utility: Gaussian process regression and Bayesian learning.
		Evaluation is done by confronting the existing agents that are able to negotiate in such context: Random Walker, Tit-for-tat and Nice Tit-for-Tat. None of those agents succeeds in beating our agent; moreover the modular and adaptive nature of our approach is a huge advantage when it comes to optimize it in specific applicative contexts.

       	\textbf{Bargaining and negotiation, Learning agent-to-agent interactions (negotiation, trust, coordination)}
	\end{abstract}

\section{Introduction}
	        
	\label{sec:introduction}
	Negotiation is a form of interaction in which a group of agents with conflicting interests and a desire to cooperate try to reach a mutually acceptable agreement on an object of negotiation \cite{Baarslag2015Learning}. The agents explore solutions according to a predetermined protocol in order to find an acceptable agreement.	
	Being widely used in economic domains and with the rise of e-commerce applications, the question of automating negotiation has gained a lot of interest in the field of artificial intelligence and multi-agent systems.
	
	Many negotiation frameworks have been proposed \cite{Guttman1998Agent} and encompass different aspects: the set of participants, the agent preferences and the number of issues. One of the major issues in automated negotiation is considering the time pressure to be well defined over the negotiation, and materialized through a deadline. However, some potential applications of automated negotiation, typically in industrial context, may require an varying time pressure. Factoring is a good example: when a company sells goods or services to another company, it produces an invoice which may be paid after several weeks. This delay of payment may have a negative impact on the company activity, as it may not have sufficient liquidity to fulfill other contracts. Factoring is an interesting answer to this issue. A funding company -- called a {\it factor} -- accepts to fund the invoices of the supplier, by paying them immediately less than their nominal amount and assuming the delay of payment of the principal.  From the factor perspective, it can be seen as a short term investment.
    This setting may include issues of various kinds:  continuous (discount rate), numeric (nominal amount of invoices), and categorical (prinicpal).In this application, time pressure is not constant over the negotiation. For the factor, the time pressure depends both on the money it has to invest and on the investment opportunities. For the supplier, the negotiation may be suddenly interrupted at some point if the payment of an invoice makes it useless for the supplier. 

	In this paper, we introduce a loosely constrained adaptive strategy for automated negotiation that can negotiate with nonlinear preferences, over discrete and continuous issues and without predefined time pressure. We represent bargaining as an extensive game with each proposal considered as a move in the game \cite{Nash1950,rubinstein1982perfect}. We state that our opponent is adaptive, which implies that negotiation history is important. We therefore propose to combine Monte Carlo Tree Search (MCTS) and opponent modeling. MCTS has proved to be a very adaptable game heuristics, in General Game Playing  \cite{Finnsson2012}. It has also proved efficient for large branching factor games \cite{Browne2012survey,Silver2016}.
	
	
	

	\section{Related works}
	\label{relatedworks}
	
    In this section, we introduce the domains related to our agent: automated negotiation and Monte Carlo methods applied to games.
	
	\paragraph{Automated negotiation}
    
To review automated negotiation strategies, we rely on the ``BOA'' (Bidding strategy, Opponent modeling, Acceptance strategy) paradigm \cite{Baarslag2016Exploring}:

Bidding strategies may depend on the history, \textsl{i.e.} the concessions made by the opponent, a negotiation deadline, the utility function of the agent, and the opponent model.
    In particular, Tit-for-tat \cite{Faratin1998Negotiation} only relies on the opponent proposals. Nice Tit-For-Tat agent \cite{Baarslag2013tit} uses learning techniques in order to improve it. The other methods rely on time pressure and cannot be applied in our context.

	Acceptance strategies can be divided into two main categories \cite{Baarslag2013}. The first category is called ``myopic strategies'' as they only consider the last bid of the opponent, the agent's own last proposal or its bidding strategy. The second category consists of ``optimal strategies'' \cite{Baarslag2013} and rely on the deadline.
  
	Most of the opponent modeling techniques related to automated negotiation have been reviewed by \cite{Baarslag2015Learning}. 
    In our context, they can be used to model: the opponent bidding strategy, its utility and the acceptance strategy. There are two main methods to model adaptive \textbf{bidding strategies} which do not rely on the deadline: neural networks and time series-based techniques. Among time series techniques, the Gaussian process regression is a stochastic technique which has been used with success by \cite{Williams2011Using}.
	The opponent \textbf{utility} is generally modeled through either frequency based techniques or Bayesian learning. Frequency methods  are relevant in the cases where the negotiation domain only consists of discrete issues. Bayesian Learning \cite{Hindriks2008Opponent} is well suited for the continuous case and can easily be extended to categorical domains.
	The opponent \textbf{acceptance strategy} can be learned in two ways, either by assuming that the opponent has a myopic strategy or by using neural networks \cite{Fang2008Opponent}. The latter is computationally expensive though.
	
	\paragraph{Monte Carlo Tree Search}
	
	Monte Carlo methods are often used as heuristics for games. 
    Kocsis and Szepevsv\'ari \cite{Kocsis2006Bandit} propose a method to combine the construction of a game tree with Monte Carlo techniques. This method is called Monte Carlo Tree Search (MCTS). It consists of 4 steps: \textbf{selection} explores the already built part of the tree, \textbf{expansion} generates a new node, \textbf{simulation} plays a game until a final state is reached and the utility of the agents is computed and \textbf{backpropagation} propagated it over all the selected nodes.
	
	\section{Our MCTS-based agent}
	Our agent follows the BOA paradigm. It consists of a bidding strategy that implements MCTS and an opponent modeling module divided into two submodules: one for the opponent utility, the other for its bidding strategy. The last module is the acceptance strategy, it accepts the opponent proposal if it is better than the bid generated by the bidding strategy.
	\label{secmocana}

\paragraph{Opponent modeling}
	\label{ch5-2-opponent-modeling}
	To model the opponent bidding strategy, we use Gaussian Process Regression \cite{Rasmussen2006Gaussian,Williams2011Using}. One of the capital aspects of this method is the choice of its kernel. In order to choose the best one, we compared them using an automated negotiation setting similar to the one of our experiments. The \cref{kernel-table} shows the results of GPR for each of the most common ones over 50 negotiation sessions. This method can also be used on the categorical issues as explained in chapter~3 of \cite{Rasmussen2006Gaussian}.
    
	\begin{table}[!ht]
		\setlength{\tabcolsep}{8pt}	
		\centering
		\begin{tabular}{|r|c| c| c| c|}
			\hline
			\textsl{Kernel} & RBF & RQF & Matérn & ESS\\\hline 
			\textsl{avg. dist.} & 43.288 & \textbf{17.766} & 43.228 & 22.292\\\hline
		\end{tabular}
		\caption{Avg dist. between proposals and GPR predictions}
		\label{kernel-table}
	\end{table}
    
    Bayesian learning \cite{Hindriks2008Opponent} considers that an agent makes concessions at roughly constant rate. It relies on triangular functions. It first generates a predetermined number of hypotheses on the utility functions and the estimate of their probability based on received proposals. The estimated utility of the opponent is the sum of these hypotheses weighted by their probability. This method can be naturally extended to the categorical issues by using traditional Bayesian inference.

    A simulated agent accepts the proposal from its opponent if its utility is better than the utility generated by its bidding strategy.

	\paragraph{MCTS-based bidding strategy}
    As explained in introduction, negotiation can be considered as a 2 players extensive game \cite{Nash1950,rubinstein1982perfect}. However, we must adapt the heuristics traditionally used for games to its peculiarities:
	\textbf{Selection} is based on progressive widening \cite{Couetoux2013Monte}: a new node is expanded if $n_p^\alpha\geq n_c$, with $n_p$ the number of simulations of the parent, $n_c$ its number of children and $\alpha$ a parameter of the model. If there is no expansion, the selected node $i$ maximizes $W_i = \frac{s_i}{n_i+1} + C\times n^\alpha\sqrt{\frac{\ln(n)}{n_i+1}}$ with $n$ the number of simulations of the tree, $s_i$ the score of the node $i$ and $C$ a parameter of the model.
    \textbf{Expansion} is chosen randomly among possible bids.
    \textbf{Simulation} is made according to the opponent models.
    \textbf{Backpropagation} is made both on the agent score and the opponent modeled score.

    We use the agent knowledge on the game to prune the less promising branches of the tree: \textsl{i.e.} all the branches of the tree with lower utility than the best proposal of the opponent (from our agent's point of view).

    \section{Results}
   	\label{experimentalAnalysis}
	 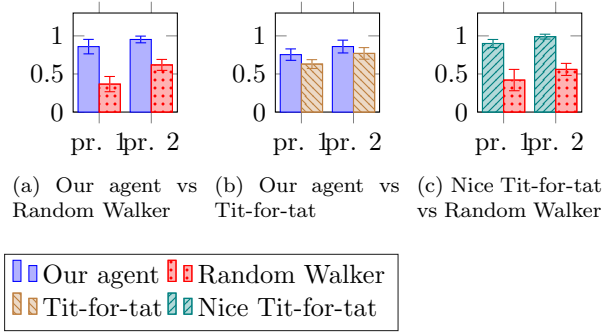
\begin{figure}[!t]
	 	\hfill
	 	\subfloat[Our agent vs Random Walker]{
	 		\begin{tikzpicture}
	 		\begin{axis}[
	 		nodes/.style={color=black},
	 		color=black,
	 		ymin=0,
	 		ymax=1.3,
	 		ybar=0pt,
	 		enlarge x limits=0.5,
	 		bar width=8,
	 		xtick={1, 2},
	 		xticklabels={pr. 1,pr. 2},
	 		width=.18\textwidth,
	 		height=.15\textheight
	 		]
	 		\addplot+[style={color=blue,fill=blue!30},error bars/.cd,
	 		y dir=both,y explicit]
	 		coordinates {
	 			(1,0.859) +- (0.0, 0.095)
	 			(2,0.953) +- (0.0, 0.044)};
	 		\addplot+[style={color=red,fill=red!30,postaction={pattern=dots,pattern color = red}},error bars/.cd,
	 		y dir=both,y explicit]
	 		coordinates {
	 			(1,0.367) +- (0.0, 0.1)
	 			(2,0.619) +- (0.0, 0.072)};
	 		\end{axis}
	 		\end{tikzpicture}
	 		\label{res-rw}
	 	}\hfill
	 	\subfloat[Our agent vs Tit-for-tat]{
	 		\begin{tikzpicture}
	 		\begin{axis}[
	 		nodes/.style={color=black},
	 		color=black,
	 		ymin=0,
	 		ymax=1.3,
	 		ybar=0pt,
	 		enlarge x limits=0.5,
	 		bar width=8,
	 		xtick={1, 2},
	 		xticklabels={pr. 1,pr. 2},
	 		width=.18\textwidth,
	 		height=.15\textheight
	 		]
	 		\addplot+[style={color=blue,fill=blue!30},error bars/.cd,
	 		y dir=both,y explicit]
	 		coordinates {
	 			(1,0.754) +- (0.0, 0.075)
	 			(2,0.86) +- (0.0, 0.084)};
	 		\addplot+[style={color=brown,fill=brown!30,postaction={pattern=north west lines,pattern color = brown}},error bars/.cd,
	 		y dir=both,y explicit]
	 		coordinates {
	 			(1,0.63) +- (0.0, 0.058)
	 			(2,0.77) +- (0.0, 0.076)};
	 		\end{axis}
	 		\label{res-tft}
	 		\end{tikzpicture}
	 	}\hfill
	 	\subfloat[Nice Tit-for-tat vs Random Walker]{
	 		\begin{tikzpicture}
	 		\begin{axis}[
	 		nodes/.style={color=black},
	 		color=black,
	 		ymin=0,
	 		ymax=1.3,
	 		ybar=0pt,
	 		enlarge x limits=0.5,
	 		bar width=8,
	 		xtick={1, 2},
	 		xticklabels={pr. 1,pr. 2},
	 		width=.18\textwidth,
	 		height=.15\textheight
	 		]
	 		\addplot+[style={color=teal,fill=teal!30,postaction={pattern=north east lines,pattern color = teal}},error bars/.cd,
	 		y dir=both,y explicit]
	 		coordinates {
	 			(1,0.90) +- (0.0, 0.053)
	 			(2,0.99) +- (0.0, 0.032)};
	 		\addplot+[style={color=red,fill=red!30,postaction={pattern=dots,pattern color = red}},error bars/.cd,
	 		y dir=both,y explicit]
	 		coordinates {
	 			(1,0.42) +- (0.0, 0.14)
	 			(2,0.56) +- (0.0, 0.079)};
	 		\end{axis}
	 		\end{tikzpicture}
	 		\label{res-ntft}
	 	}\hfill
	 	\subfloat{
             \centering
	 		\begin{tikzpicture} 
	 		\begin{axis}[%
	 		width=.2\textwidth,
	 		height=20mm,
	 		hide axis,
	 		xmin=10,
	 		xmax=50,
	 		ymin=0,
	 		ymax=0.4,
	 		ybar,
	 		legend style={draw=white!15!black,legend cell align=left},
	 		legend columns=2
	 		]
	 		\addlegendimage{color=blue,fill=blue!30}
	 		\addlegendentry{Our agent};
	 		\addlegendimage{color=red,fill=red!30,postaction={pattern=dots,pattern color = red}}
	 		\addlegendentry{Random Walker};
	 		\addlegendimage{color=brown,fill=brown!30,postaction={pattern=north west lines,pattern color = brown}}
	 		\addlegendentry{Tit-for-tat};
	 		\addlegendimage{color=teal,fill=teal!30,postaction={pattern=north east lines,pattern color = teal}}
	 		\addlegendentry{Nice Tit-for-tat};
	 		\end{axis}
	 		\end{tikzpicture}
	 	}
	 	\hfill
	 	\caption{Average utility of negotiating agents}
	 	\label{results}
	 \end{figure}
	 
    Our agent is evaluated using the \textsc{Genius} \cite{Lin2014} framework against Tit-for-tat \cite{Faratin1998Negotiation}, Nice Tit-for-Tat and RandomWalker \cite{Baarslag2013tit} in the ANAC 2014 setting \cite{Fukuta2016} except that there is no deadline.
    \Cref{results} displays the utility of the agents when negotiating with each other. The utility of the agents is displayed as an histogram. The results are averaged over 20 negotiation sessions with each profile, with error bars representing the standard deviation from the average.
    
	Our agent is able to beat the Random Walker in every situation and get a significantly better result whatever the profile. Our agent gets a lower utility with Tit-for-Tat but is still able to beat it significantly. The negotiations with Nice Tit-for-Tat never end: the agents keep negotiating forever. We propose instead an indirect evaluation by confronting Nice Tit-for-Tat with Random Walker, in the same setting. The performances of both agents are equal, considering the standard deviation of the series.
    
	\section{Conclusion}
	\label{conclusion}
	In this paper, we presented a negotiating agent able to negotiate in a context where agents do not have predetermined deadline, with both continuous and categorical issues. The experimental results are promising: against all the agents that can negotiate in this domain, our agent outperformed Random Walker and Tit-for-Tat and draws with Nice Tit-for-Tat. Among the perspectives of this work, we propose to adapt it to the multilateral context and try improvements as AMAF and RAVE \cite{AMAF, Rave}
	\bibliographystyle{plain}
	\balance
	\bibliography{biblio.bib}

\begin{thebibliography}{10}

\bibitem{Baarslag2016Exploring}
Tim Baarslag.
\newblock {\em Exploring the Strategy Space of Negotiating Agents: A Framework
  for Bidding, Learning and Accepting in Automated Negotiation}.
\newblock PhD thesis, Delft University of Technology, 2016.

\bibitem{Baarslag2015Learning}
Tim Baarslag, Mark J~C Hendrikx, Koen~V Hindriks, and Catholijn~M Jonker.
\newblock Learning about the opponent in automated bilateral negotiation: a
  comprehensive survey of opponent modeling techniques.
\newblock {\em Autonomous Agents and Multi-Agent Systems}, 20(1):1--50, 2015.

\bibitem{Baarslag2013}
Tim Baarslag and Koen~V. Hindriks.
\newblock Accepting optimally in automated negotiation with incomplete
  information.
\newblock In {\em AAMAS '13}, pages 715--722, Richland, SC, 2013. International
  Foundation for Autonomous Agents and Multiagent Systems.

\bibitem{Baarslag2013tit}
Tim Baarslag, Koen~V Hindriks, and Catholijn Jonker.
\newblock A tit for tat negotiation strategy for real-time bilateral
  negotiation.
\newblock In {\em Complex Automated Negotiations: Theories, Models, and
  Software Competitions}, volume 435, pages 229--233. Springer Berlin
  Heidelberg, 2013.

\bibitem{Browne2012survey}
Cameron~C Browne, Edward Powley, Daniel Whitehouse, Simon~M. Lucas, Peter~I
  Cowling, Philipp Rohlfshagen, Stephen Taverner, Diego Perez, Spyridon
  Samothrakis, and Simon Colton.
\newblock A survey of {M}onte {C}arlo tree search methods.
\newblock {\em IEEE Transactions on Computational Intelligence and AI in
  games}, 4(1):1--43, 2012.

\bibitem{Couetoux2013Monte}
Adrien Couëtoux.
\newblock {\em {M}onte {C}arlo Tree Search for Continuous and Stochastic
  Sequential Decision Making Problems}.
\newblock PhD thesis, Université Paris XI, 2013.

\bibitem{Fang2008Opponent}
Fang Fang, Ye~Xin, Yun Xia, and Xu~Haitao.
\newblock An opponent's negotiation behavior model to facilitate buyer-seller
  negotiations in supply chain management.
\newblock In {\em 2008 International Symposium on Electronic Commerce and
  Security}, 2008.

\bibitem{Faratin1998Negotiation}
Peyman Faratin, Nicholas~R Jennings, and Carles Sierra.
\newblock Negotiation decision functions for autonomous agents.
\newblock {\em Robotics and Autonomous Systems}, 24(3-4):159--182, 1998.

\bibitem{Finnsson2012}
Hilmar Finnsson.
\newblock Generalized {M}onte {C}arlo tree search extensions for general game
  playing.
\newblock In {\em Proceedings of the Twenty-Sixth AAAI Conference on Artificial
  Intelligence}, AAAI'12, pages 1550--1556. AAAI Press, 2012.

\bibitem{Fukuta2016}
Naoki Fukuta, Takayuki Ito, Minjie Zhang, Katsuhide Fujita, and Valentin Robu,
  editors.
\newblock {\em Recent Advances in Agent-based Complex Automated Negotiation},
  volume 638 of {\em Studies in Computational Intelligence}.
\newblock Springer International Publishing, 2016.

\bibitem{Rave}
Sylvain Gelly and David Silver.
\newblock Monte-carlo tree search and rapid action value estimation in computer
  go.
\newblock {\em Artificial Intelligence}, 175(11):1856--1875, 2011.

\bibitem{Guttman1998Agent}
Robert~H Guttman, Alexandros~G Moukas, and Pattie Maes.
\newblock Agent-mediated electronic commerce: a survey.
\newblock {\em The Knowledge Engineering Review}, 13(02):147--159, Jul 1998.

\bibitem{AMAF}
David~P Helmbold and Aleatha Parker-Wood.
\newblock All-moves-as-first heuristics in monte-carlo go.
\newblock In {\em IC-AI}, pages 605--610, 2009.

\bibitem{Hindriks2008Opponent}
Koen Hindriks and Dmytro Tykhonov.
\newblock Opponent modelling in automated multi-issue negotiation using
  bayesian learning.
\newblock In {\em Proceedings of the 7th International Joint Conference on
  Autonomous Agents and Multiagent Systems}, volume~1, pages 331--338, 2008.

\bibitem{Kocsis2006Bandit}
Levente Kocsis and Csaba Szepesv{\'a}ri.
\newblock Bandit based {M}onte {C}arlo planning.
\newblock In Johannes F{\"u}rnkranz, Tobias Scheffer, and Myra Spiliopoulou,
  editors, {\em Machine Learning: ECML 2006: 17th European Conference on
  Machine Learning Berlin, Germany, September 18-22, 2006 Proceedings}, pages
  282--293. Springer Berlin Heidelberg, 2006.

\bibitem{Lin2014}
Raz Lin, Sarit Kraus, Tim Baarslag, Dmytro Tykhonov, Koen Hindriks, and
  Catholijn~M Jonker.
\newblock Genius: an integrated environment for supporting the design of
  generic automated negotiators.
\newblock {\em Computational Intelligence}, 30(1):48--70, 2014.

\bibitem{Nash1950}
John~F Nash~Jr.
\newblock The bargaining problem.
\newblock {\em Econometrica: Journal of the Econometric Society}, pages
  155--162, 1950.

\bibitem{Rasmussen2006Gaussian}
Carl~E Rasmussen and Christopher K~I Williams.
\newblock {\em Gaussian processes for machine learning}.
\newblock MIT Press, 2006.

\bibitem{rubinstein1982perfect}
Ariel Rubinstein.
\newblock Perfect equilibrium in a bargaining model.
\newblock {\em Econometrica: Journal of the Econometric Society}, pages
  97--109, 1982.

\bibitem{Silver2016}
David Silver, Aja Huang, Chris~J Maddison, Arthur Guez, Laurent Sifre, George
  Van Den~Driessche, Julian Schrittwieser, and al.
\newblock Mastering the game of go with deep neural networks and tree search.
\newblock {\em Nature}, 529(7587):484--489, 2016.

\bibitem{Williams2011Using}
Colin~R Williams, Valentin Robu, Enrico~H Gerding, and Nicholas~R Jennings.
\newblock Using gaussian processes to optimise concession in complex
  negotiations against unknown opponents.
\newblock In {\em IJCAI'11}, pages 432--438, 2011.

\end{thebibliography}
\end{document}